\documentclass[10pt,twocolumn,letterpaper]{article}

\usepackage{cvpr}              
\usepackage{amsfonts,amssymb}
\usepackage{multirow}
\usepackage{graphicx}
\usepackage{tabularx}
\usepackage{lipsum}
\usepackage{multicol}
\usepackage[accsupp]{axessibility}
\renewcommand{\thefootnote}{}

\usepackage[dvipsnames]{xcolor}

\definecolor{cvprblue}{rgb}{0.21,0.49,0.74}
\usepackage[pagebackref,breaklinks,colorlinks,citecolor=cvprblue]{hyperref}

\usepackage{enumitem}
\setlist{itemsep=0pt,parsep=0pt}

\title{AMU-Tuning: Effective Logit Bias for CLIP-based Few-shot Learning}

\author{Yuwei Tang$^{*}$, Zhenyi Lin$^{*}$, Qilong Wang$^{\dagger}$, Pengfei Zhu, Qinghua Hu\\Tianjin Key Lab of Machine Learning, College of Intelligence and Computing, Tianjin University, China\\
\texttt{\{tangyuwei, linzhenyi, qlwang, zhupengfei, huqinghua\}@tju.edu.cn}\\
}

\begin{document}
\maketitle

\begin{abstract}
Recently, pre-trained vision-language models (e.g., CLIP) have shown great potential in few-shot learning and attracted a lot of research interest. Although efforts have been made to improve few-shot ability of CLIP, key factors on the effectiveness of existing methods have not been well studied, limiting further exploration of CLIP's potential in few-shot learning. In this paper, we first introduce a unified formulation to analyze CLIP-based few-shot learning methods from a perspective of logit bias, which encourages us to learn an effective logit bias for further improving performance of CLIP-based few-shot learning methods. To this end, we disassemble three key components involved in computation of logit bias (i.e., logit features, logit predictor, and logit fusion) and empirically analyze the effect on performance of few-shot classification. Based on analysis of key components, this paper proposes a novel AMU-Tuning method to learn effective logit bias for CLIP-based few-shot classification. Specifically, our AMU-Tuning predicts logit bias by exploiting the appropriate \underline{\textbf{A}}uxiliary features, which are fed into an efficient feature-initialized linear classifier with \underline{\textbf{M}}ulti-branch training. Finally, an \underline{\textbf{U}}ncertainty-based fusion is developed to incorporate logit bias into CLIP for few-shot classification. The experiments are conducted on several widely used benchmarks, and the results show AMU-Tuning clearly outperforms its counterparts while achieving state-of-the-art performance of CLIP-based few-shot learning without bells and whistles.
\footnote{$^*$ Equal contributions made by Y. Tang and Z. Lin, \ $\dagger$ Corresponding author is Q. Wang. This work was supported in part by National Natural Science Foundation of China under Grants 62276186, 61925602, 62222608, in part by CAAI-Huawei MindSpore Open Fund under Grant CAAIXSJLJJ-2022-010 C, in part by Tianjin Natural Science Funds for Distinguished Young Scholar under Grant 23JCJQJC00270, and in part by the Haihe Lab of ITAI under Grant 22HHXCJC00002.}
\end{abstract} 
\section{Introduction}
\label{sec:intro}
\begin{figure}
    \centering
    \includegraphics[width=0.45\textwidth]{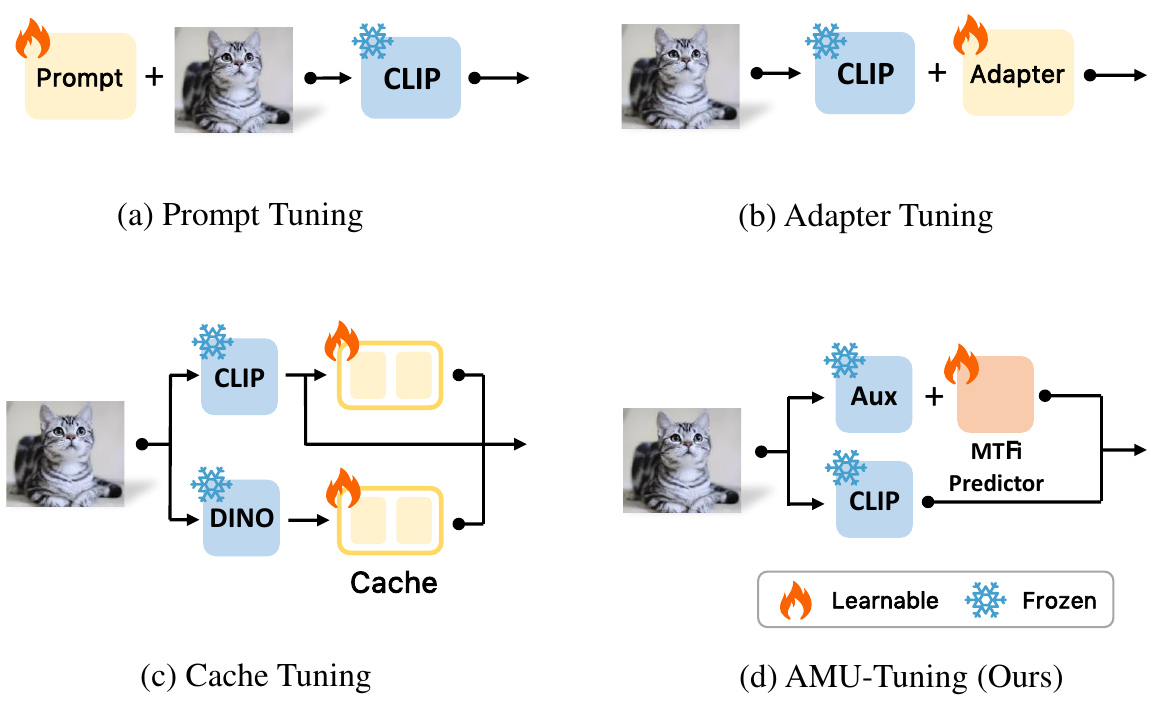}
    \caption{Comparison of the existing CLIP-based few-shot learning methods in terms of architecture design.}
    \label{fig:sa}
\end{figure}

In recent years, large-scale vision-language models~\cite{clip,coca,flamingo,beitv3,pali,gpt4,jin2023context}  have attracted large amounts of research attention in computer vision community (especially CLIP~\cite{clip}), due to the remarkable performance on downstream tasks, e.g., zero-shot generalization ability~\cite{calip,sus,yang2023zero}. Recently, some efforts have been made to improve the transfer learning ability of CLIP given a limited set of training samples~\cite{coop,cocoop,clip-adapter,tip,cafo}, i.e., CLIP-based few-shot learning. These methods can be roughly divided into three categories: (1) prompt tuning~\cite{coop,cocoop}; (2) adapter-based tuning~\cite{clip-adapter,tip}; (3) cache-based tuning~\cite{tip,cafo}. Specifically, as illustrated in \cref{fig:sa}, prompt-tuning methods improve the few-shot learning ability of CLIP by introducing learnable text prompt~\cite{coop,cocoop} for the text encoder of CLIP. For adapter-based tuning, some lightweight modules, e.g., multi-layer perceptron (MLP)~\cite{clip-adapter}, are built at the end of text and visual encoders to adjust text and visual features for downstream tasks. Subsequently, cache-based tuning methods~\cite{tip,cafo} present ``soft" $K$-nearest neighbor classifiers storing visual features and labels of training samples, which are combined with zero-shot CLIP for final classification.

\begin{figure*}
    \centering
    \includegraphics[width=0.95\textwidth]{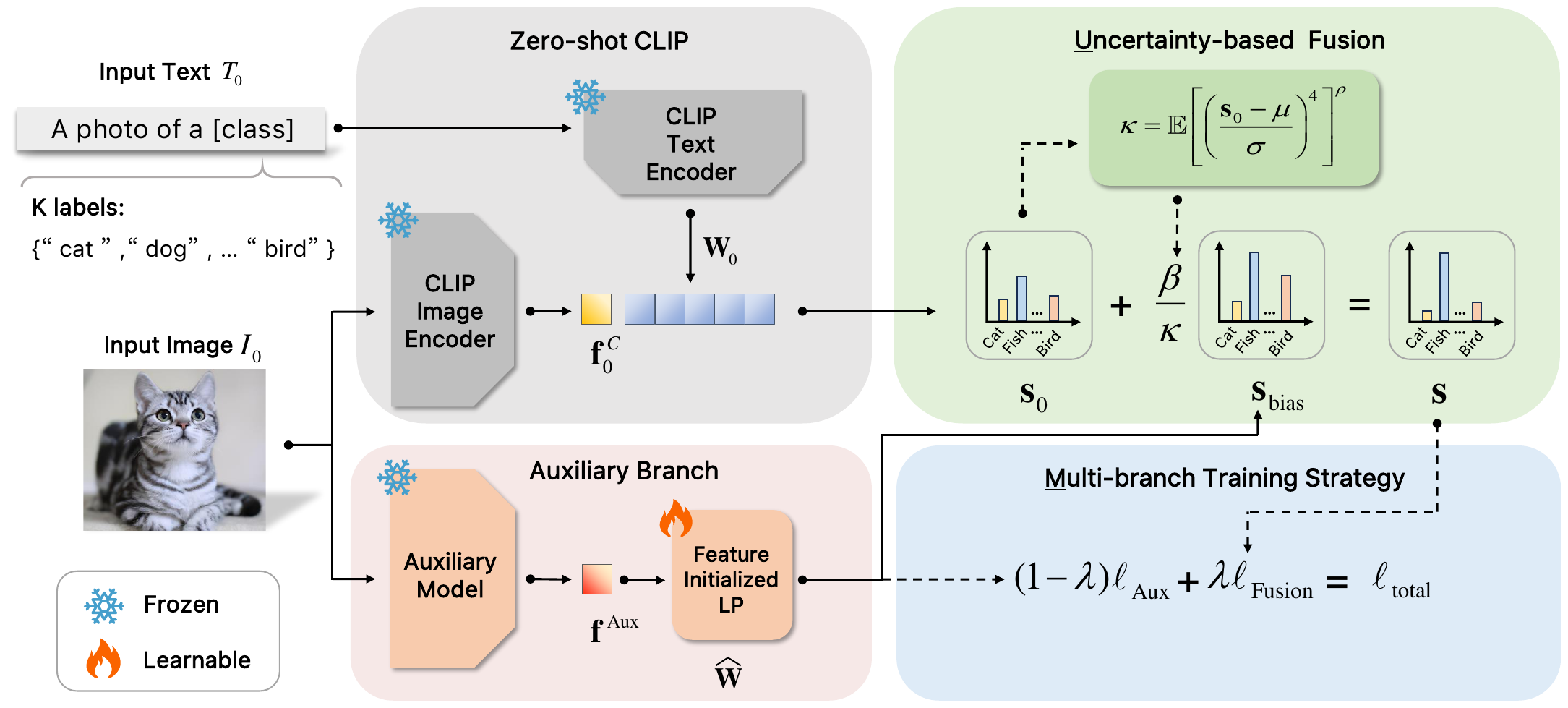}
    \caption{Overview of our proposed AMU-Tuning method for CLIP-based few-shot classification. Specifically, our AMU-Tuning exploits the complementary \underline{\textbf{A}}uxiliary features to compute logit bias. Then, an efficient feature-initialized LP with \underline{\textbf{M}}ulti-branch training is presented to improve performance of logit predictor by better exploring the auxiliary features. Finally, we develop a \underline{\textbf{U}}ncertainty-based fusion by considering prediction confidence of zero-shot CLIP, which adaptively incorporates logit bias into CLIP for few-shot classification.}
    \label{fig:mainfig}
\end{figure*}

Although many works have been studied to improve the few-shot generalization ability of CLIP, the relationship among the existing methods seems a bit loose. More importantly, the key factors on the effectiveness of existing methods have not been well studied, which limits further exploring the potential of CLIP to few-shot learning. Therefore, this paper introduces a unified formulation to analyze CLIP-based few-shot learning methods from a perspective of logit bias, where we show most of the previous methods can be generally regarded as learning different logit biases for zero-shot CLIP. Meanwhile, logit bias dramatically impacts the performance of few-shot classification. It encourages us to learn an effective logit bias for further improving performance of CLIP-based few-shot learning methods. 

According to the observations on previous methods from the perspective of logit bias, we disassemble three key components involved in logit bias (i.e., logit features, logit predictor, and logit fusion), while empirically analyzing the effect on few-shot classification. Specifically, we first compare several auxiliary features to predict logit bias in terms of complementary and superiority, while showing the appropriate features greatly help to learn effective logit bias. Then, we evaluate the effect of various logit predictors, e.g., MLP, cache-based model, and linear probing (LP), while showing feature initialization is helpful for logit predictor, but existing logit predictors do not fully explore the superiority of auxiliary features. Finally, we observe that trade-off parameter of fusion is very sensitive to models and datasets, which is related to prediction confidence of zero-shot CLIP.  

Based on above analysis on the key components, we propose a novel AMU-Tuning method to learn effective logit bias for CLIP-based few-shot classification. Specifically, our AMU-Tuning exploits a kind of \underline{\textbf{A}}uxiliary features complementary to CLIP for computing logit bias. Then, an efficient feature-initialized LP with \underline{\textbf{M}}ulti-branch training is presented to improve the performance of logit predictor by better exploring the potential of auxiliary features. Finally, we develop a \underline{\textbf{U}}ncertainty-based fusion by considering prediction confidence of zero-shot CLIP, which adaptively incorporates logit bias into CLIP for effective few-shot classification. The overview of our AMU-Tuning is shown in \cref{fig:mainfig}. To evaluate the effectiveness of our AMU-Tuning method, experiments are conducted on eleven downstream tasks~\cite{caltech101,car,ucf101,flower102,sun397,dtd,eurosat,fvga,pet,food101,imagenet}, and four out-of-distribution benchmarks~\cite{imagenet-a,imagenet-r,imagenet-s,imagenet-v2} by using various backbone models (i.e., ResNets~\cite{resnet} and ViT~\cite{vit}). The contributions of this work are summarized as follows:
\begin{itemize}
    \item To our best knowledge, this work makes the first attempt to introduce a unified formulation for CLIP-based few-shot learning methods from a perspective of logit bias. It allows us to further explore the effectiveness of existing methods by analyzing the effect of three key components involved in logit bias, i.e., features, predictor, and fusion.

    \item Based on the analysis on the key components of logit bias, we propose an efficient AMU-Tuning method for CLIP-based few-shot classification, whose core is to learn effective logit bias by exploiting the appropriate \underline{\textbf{A}}uxiliary features with \underline{\textbf{M}}ulti-branch training of a feature-initialized linear classifier, followed by an \underline{\textbf{U}}ncertainty-based fusion.  

    \item Extensive experiments are conducted on several downstream tasks and out-of-distribution benchmarks, and the results show our proposed AMU-Tuning clearly outperforms its counterparts while achieving state-of-the-art performance of CLIP-based few-shot learning with more efficient computational cost.
\end{itemize}

\begin{table*}[htbp]
\renewcommand{\arraystretch}{1.35}
\tabcolsep=4pt
\footnotesize
\centering
\begin{tabular}{llcccc}
\toprule
Model                                                  & Bias & Feature & Predictor & Fusion & 16-shot Acc (\%) \\ \midrule
Zero-shot CLIP~\cite{clip} &  -    &   -      &       -    &    -    &   60.33  \\
CoOp~\cite{coop} & $\simeq f_{T}(T_{\text{bias}})\mathbf{f}_0^C$ & $T_{\text{bias}}$ & $f_{T}$  & -&   62.95  \\
CLIP-Adapter~\cite{clip-adapter} &   $f_T^{\text{Ada}}(\mathbf{W}_0)\mathbf{f}_0^C+\mathbf{W}_0f_V^{\text{Ada}}(\mathbf{f}_0^C)+f_T^{\text{Ada}}(\mathbf{W}_0)f_V^{\text{Ada}}(\mathbf{f}_0^C)$   & CLIP  & MLP & Manual Tuning &   63.59  \\
Tip-Adapter-F~\cite{tip}& $\phi\left(\mathbf{F}_{\text{TrC}}^{\mathsf{T}}\mathbf{f}_{0}^{C}\right)\mathbf{V}$ & CLIP & Cache & Manual Tuning &  65.51   \\
CaFo~\cite{cafo} & $\alpha\phi\left(\mathbf{F}_{\text{TrC}}^{\mathsf{T}}\mathbf{f}_{0}^{C}\right)\mathbf{V} + (1-\alpha)\phi\left(\mathbf{F}_{\text{TrD}}^{\mathsf{T}}\mathbf{f}_{0}^{D}\right)\mathbf{V}$ & CLIP+DINO & Cache & Similarity-based &   68.79  \\
AMU-Tuning (Ours) & $\widehat{\mathbf{W}}\mathbf{f}^{\text{Aux}}$ & Aux & MTFi LP & Uncertainty-based  &  \textbf{70.02}  \\ \bottomrule
    \end{tabular}%
    \caption{Comparison of existing CLIP-based few-shot learning methods from the perspective of logit bias. Different from previous works, our AMU-Tuning learns logit bias by exploiting the appropriate auxiliary (Aux) features with multi-branch training feature-initialized (MTFi) LP followed by an uncertainty-based fusion, while achieving higher accuracy (Acc) on ImageNet-1K with 16-shot training samples.}
  \label{tab:formulbias}%
\end{table*}%

\section{Related Work}
\label{sec:formatting}
\paragraph{Few-shot Classification}
Few-shot learning is crucial for limited-sample scenarios, with extensive exploration in this field~\cite{o1,o2,o3,o4,o5,o6,o7,o8}. Recently, there's been a surge in few-shot learning methods that fine-tune large pre-trained models while CLIP~\cite{clip} makes a seminal work to train a large-scale vision-language model, showing good generalization on various downstream tasks. Subsequently, several works have been studied to improve few-shot generalization ability of CLIP. CoOp~\cite{coop} first shows text prompt greatly impacts on zero-shot performance of CLIP and introduces the idea of prompt learning to improve few-shot performance of CLIP. After that, a lot of works are proposed to improve the effectiveness of prompt learning~\cite{coop,cocoop,upl,plot}. However, prompt learning methods need to compute the gradients throughout text encoder, suffering from the heavy computational cost. Consequently, CLIP-Adapter~\cite{clip-adapter} propose a residual structure and MLP-based adapter modules to fine-tune outputs of text and visual encoders. Tip-Adapter~\cite{tip} presents a cache structure to perform a ``soft" $K$-nearest neighbor classifier, which is combined with zero-shot CLIP. Going beyond Tip-Adapter, CaFo~\cite{cafo} introduces an extra cache structure with DINO~\cite{dino}, while employing  GPT-3~\cite{gpt3} and DALL-E~\cite{dalle} models for text and visual information augmentation. Different from above works, our AMU-Tuning learns logit bias by exploiting the appropriate auxiliary features with multi-branch training feature-initialized LP followed by an uncertainty-based fusion, while achieving better performance.


\vspace{-.65cm}
\paragraph{Large-scale Pre-trained Models} Inspired by the success of large-scale pre-trained models in the field of natural language processing~\cite{bert,T5,gpt3,gpt4}, many researchers have undertaken various explorations in pre-training visual models in the domain of computer vision, including ResNet~\cite{resnet}, ViT~\cite{vit}, Swin Transformers~\cite{swin}, and others~\cite{deit,swinv2,videomae}. In these large-scale pre-trained models, large-scale vision-language models represented by CLIP~\cite{clip}, SLIP~\cite{slip}, and CoCa~\cite{coca} have explored training on massive data of images and text from the internet. These models showcase powerful generalization abilities and have achieved remarkable performance in downstream tasks. Recently, a class of methods~\cite{mocov1,mocov2,mocov3,cmc,wmc} based on self-supervised learning has garnered significant attention for substantially improving the transfer performance of visual models. For instance, MoCo~\cite{mocov1} introduced a momentum-contrast framework, which reduced the constraints on batch size during training, leading to enhanced model transferability. Following that, a class of methods based on Masked Image Modeling (MIM) was devised. These methods involve masking a portion of an image and training the model to reconstruct these masked regions, achieving notable generalization performance. For example, MAE~\cite{mae} achieves pixel reconstruction through masking, while BEiT~\cite{beitv1} enhances model performance by designing a pretext task for visual tokens. Numerous other MIM-based approaches have been proposed~\cite{ibot,milan,spark,beitv3}, significantly advancing the improvement of model transferability. In our work, we aim to leverage large-scale pre-trained models as auxiliary features to collaboratively enhance few-shot generalization of CLIP.


\section{Proposed Method}
In this section, we first summarize and compare the existing methods from a perspective of logit bias. Then, we empirically analyze three key components involved in computation of logit bias. Based on the analysis, we propose an efficient and effective AMU-Tuning method.

\subsection{Perspective of Logit Bias for CLIP-based Few-shot Learning}\label{sec_3.1}
To analyze the existing methods in a unified framework, we summarize them from a perspective of logit bias. As listed in \cref{tab:formulbias}, we formulate previous works~\cite{coop,clip-adapter,tip,cafo} as
\vspace*{-0.5\baselineskip}
\begin{align}\label{eq:1}
    \mathbf{s} \cong \mathbf{W}_0\mathbf{f}_0^C + \beta\cdot\mathbf{s}_{\text{bias}}, 
    \vspace*{-0.5\baselineskip}
\end{align}
where $\mathbf{W}_0=f_{T}(T_{0})$ and  $\mathbf{f}_0^C=f_{V}(I_{0})$ indicate the outputs of text encoder $f_{T}$ and visual encoder $f_{V}$ of CLIP~\cite{clip}, respectively. $T_{0}$ and ${I_{0}}$ are text and visual inputs of zero-shot CLIP, respectively. $\mathbf{s}_0 = \mathbf{W}_0\mathbf{f}_0^C$ represents the prediction logit of zero-shot CLIP. $\mathbf{s}_{\text{bias}}$ means the logit bias learned by few-shot training samples, which is fused with $\mathbf{s}_0$ to obtain the final prediction $\mathbf{s}$. $\beta$ is a hyper-parameter of fusion. 

\begin{table}[t]
  \centering
    \begin{tabular}{lccc}
    \toprule
    Model & $\text{SUP}_{\text{Aux}}$ (\%)     & $\text{CMY}_{\text{Aux}}$  & Fusion (\%) \\
    \midrule
    ZS-CLIP~\cite{clip} & N/A  & N/A  & 60.33  \\
    CLIP~\cite{clip}  & 56.93  & 0.438  & 65.34  \\
    DINO~\cite{dino}  & 55.65  & \underline{0.816}  & 68.32 \\
    MoCov3~\cite{mocov3}  & \underline{57.68}  & \textbf{0.837}  & \textbf{69.35} \\
    MAE~\cite{mae}   & 38.98  & 0.722  & 65.49  \\
    SparK~\cite{spark} & 28.31  & 0.770  & 63.56  \\
    MILAN~\cite{milan} & \textbf{66.36}  & 0.718  & \underline{69.24}  \\
    \bottomrule
    \end{tabular}%
  \caption{Comparison of different auxiliary features in terms of complementary ($\text{CMY}_{\text{Aux}}$), superiority ($\text{SUP}_{\text{Aux}}$) and fused results on ImageNet-1K with 16-shot samples. The best and second-best results are highlighted in \textbf{bold} and \underline{underline}, respectively.}
  \label{tab:SelectModel}%
\end{table}%

Specifically, from~\cref{tab:formulbias} we can see that prompt tuning~\cite{coop,cocoop} can be approximated by combining zero-shot CLIP (i.e., $\mathbf{s}_0$) with $f_{T}(T_{\text{bias}})\mathbf{f}_0^C$, where $\mathbf{s}_{\text{bias}}$ is computed based on the learnable text prompts (i.e., $T_{\text{bias}}$). Clearly, these prompt-tuning methods require to compute the gradients throughout the text encoder $f_{T}$, suffering from the heavy computational cost. CLIP-Adapter~\cite{clip-adapter} calculates $\mathbf{s}_{\text{bias}}$ by
 \vspace{-2mm}
\begin{align}\label{eq:2}
f_T^{\text{Ada}}(\mathbf{W}_0)\mathbf{f}_0^C+\mathbf{W}_0f_V^{\text{Ada}}(\mathbf{f}_0^C)+f_T^{\text{Ada}}(\mathbf{W}_0)f_V^{\text{Ada}}(\mathbf{f}_0^{C}),
\end{align}
where $f_T^{\text{Ada}}(\cdot)$ and $f_V^{\text{Ada}}(\cdot)$ are adapters for text and visual features, which are achieved by two MLP. Cache-based Tip-Adapter~\cite{tip} and CaFo~\cite{cafo} perform a ``soft" $K$-nearest neighbor classifier on a trainable cache of visual CLIP features ($\mathbf{F}_{\text{TrC}}$) to generate $\mathbf{s}_{\text{bias}}$, i.e.  \vspace{-2mm}
\begin{align}\label{eq:3}
\phi\left(\mathbf{F}_{\text{TrC}}^{\mathsf{T}}\mathbf{f}_{0}^{C}\right)\mathbf{V},
\end{align}
where $\phi(\cdot)$ is a non-linear function, and $\mathbf{V}$ is the label matrix. $\mathsf{T}$ indicates the matrix transposition. Besides, CaFo exploits an extra trainable cache of visual DINO features~\cite{dino} ($\mathbf{F}_{\text{TrD}}$) to compute $\mathbf{s}_{\text{bias}}$ as  \vspace{-2mm}
\begin{align}\label{eq:4}
\alpha\phi\left(\mathbf{F}_{\text{TrC}}^{\mathsf{T}}\mathbf{f}_{0}^{C}\right)\mathbf{V} + (1-\alpha)\phi\left(\mathbf{F}_{\text{TrD}}^{\mathsf{T}}\mathbf{f}_{0}^{D}\right)\mathbf{V},
\end{align}
where $\alpha$ is a trade-off parameter computed based on the similarity with $\mathbf{s}_0$. The analysis above shows that most of the existing methods can be regarded as learning different logit biases to adjust the prediction of zero-shot CLIP by using few-shot training samples. 

\subsection{Analysis on Computation of Logit Bias}
By regarding the performance of previous works (the last column of~\cref{tab:formulbias}), all methods are superior to zero-shot CLIP by introducing logit bias. Particularly, CLIP-Adapter~\cite{clip-adapter} performs better than prompt tuning~\cite{coop}, while cache-based methods~\cite{tip,cafo} significantly outperform CLIP-Adapter. These comparisons clearly demonstrate that different logit biases greatly influence the performance of few-shot classification. By observing the existing methods in ~\cref{tab:formulbias}, we disassemble the computation of logit bias into three key components, i.e., logit features, logit predictor, and logit fusion. To deeply analyze the effect of logit bias on the performance of few-shot classification, we conduct comprehensive empirical comparisons on three key components in the following.

\subsubsection{Features for Computation of Logit Bias}
\label{Sec_Log_Fea}
Existing works mainly exploit CLIP itself to compute logit bias. Recently, CaFo~\cite{cafo} proposes to use the extra features~\cite{dino,dalle,gpt3} to help generating logit bias, which achieves remarkable performance gain. In this paper, we call such extra features as auxiliary features. To evaluate the effect of auxiliary features, this paper conducts an empirical comparison on several kinds of auxiliary features in terms of complementary and superiority. Specifically, we conduct experiments on ImageNet-1K~\cite{imagenet} by using CLIP with the backbone of ResNet-50 (RN50) and 16-shot training samples. For auxiliary features, we compare six pre-trained models, i.e., the visual encoder of CLIP, DINO~\cite{dino}, MoCov3~\cite{mocov3}, MAE~\cite{mae}, SparK~\cite{spark}, and MILAN~\cite{milan}. The backbone for all pre-trained models is RN50, except MAE and MILAN. We utilize VIT-B/16~\cite{vit} models for MAE and MILAN, due to the unavailability of pre-trained RN50.

To compute the logit bias, we train a simple LP for all auxiliary features. Then, the logit bias is combined with prediction of zero-shot CLIP by summation for few-shot classification. Particularly, we train an individual LP for all auxiliary features within 50 epochs, whose results represent the superiority of different auxiliary features (indicated by $\text{SUP}_{\text{Aux}}$). For measuring the complementarity ($\text{CMY}_{\text{Aux}}$) of different auxiliary features, we define $\text{CMY}_{\text{Aux}}$ by inverse of similarity between LP prediction of auxiliary features ($\mathbf{s}_{\text{Aux}}$) and prediction of zero-shot CLIP ($\mathbf{s}_{0}$): \vspace{-2mm}
\begin{align}\label{eq:5}
\text{CMY}_{\text{Aux}} & = 1-\text{SIM}(\mathbf{s}_{0},\mathbf{s}_{\text{Aux}}), \nonumber\\  
\text{SIM}(\mathbf{s}_{0},\mathbf{s}_{\text{Aux}}) & = \frac{\mathbf{s}_{0}\cdot\mathbf{s}_{\text{Aux}}}{\|\mathbf{s}_{0}\|_{2}\cdot \|\mathbf{s}_{\text{Aux}}\|_{2}},
\end{align}
where $\text{SIM}$ computes the cosine similarity between $\mathbf{s}_{\text{Aux}}$ and $\mathbf{s}_{0}$. Clearly, smaller similarity means less correlation between $\mathbf{s}_{\text{Aux}}$ and $\mathbf{s}_{0}$, indicating the auxiliary features may be more complementary to zero-shot CLIP. 

As compared in \cref{tab:SelectModel}, we can see that all auxiliary features contribute performance gains to zero-shot CLIP (ZS-CLIP). In particular, we observe that complementarity ($\text{CMY}_{\text{Aux}}$) is more important than superiority ($\text{SUP}_{\text{Aux}}$) for auxiliary features. For example, CLIP is superior to DINO, but the result of fusion with DINO is much better than one of fusing CLIP, because DINO has much higher $\text{CMY}_{\text{Aux}}$. Besides, fusion of MAE achieves similar result with fusion of CLIP, although MAE has much lower $\text{SUP}_{\text{Aux}}$. Additionally, MILAN is superior to MoCov3, but fusion of MoCov3 achieves better final results due to its higher $\text{CMY}_{\text{Aux}}$. Another observation is that the auxiliary features with higher $\text{SUP}_{\text{Aux}}$ achieve better fusion results, when they have similar $\text{CMY}_{\text{Aux}}$. For example, DINO and MoCov3 have similar $\text{CMY}_{\text{Aux}}$, and MoCov3 outperforms DINO in terms of fusion result, as MoCov3 has higher $\text{SUP}_{\text{Aux}}$ than DINO. Based on the above results, we conclude that \textit{the auxiliary features with good complementarity and superiority can greatly help to compute effective logit bias for better performance. Particularly, complementarity is more important than superiority for auxiliary features}. 

\begin{table}[t]
  \centering
    \begin{tabular}{lcccc}
    \toprule
    Auxiliary Features & Individual &Joint  & Joint+ZO \\
    \midrule
    CLIP~\cite{clip}  & 56.93 &11.23& 65.34 \\
    DINO~\cite{dino}  & 55.65 & 36.24 &68.32\\
    MoCov3~\cite{mocov3}  &57.68 & 42.82 &69.35   \\
    \bottomrule
    \end{tabular}
\caption{Comparison (\%) of two training strategies (i.e., Individual and Joint) for the bias branch on ImageNet-1K. Joint+ZO indicates the fused results of joint training bias branch with zero-shot CLIP.}
\label{tab:adaptertrain}
\end{table}

\begin{figure}
    \centering
    \includegraphics[width=0.33\textwidth]{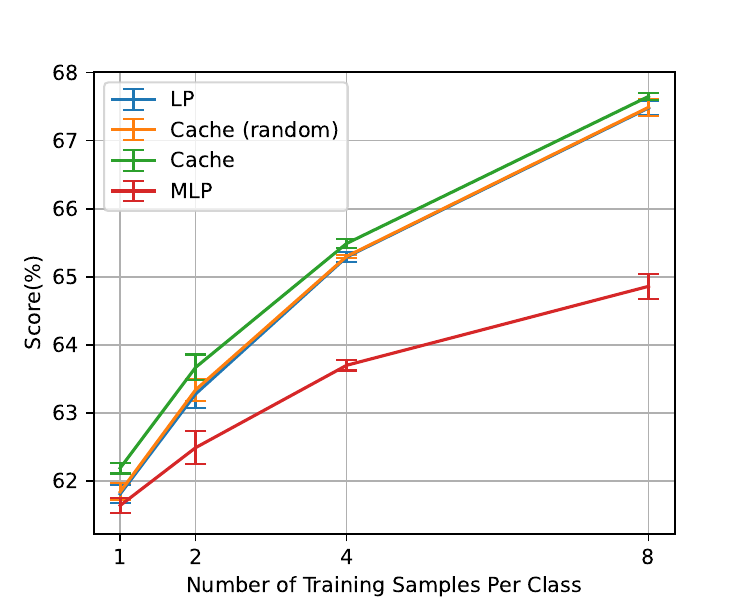}
    \caption{Results of different logit predictors on ImageNet-1K.}
    \label{fig:adapter}
\end{figure}

\subsubsection{Logit Predictor}\label{sec_log_pre}

As discussed in ~\cref{sec_3.1} and \cref{tab:formulbias}, existing works mainly exploit MLP~\cite{clip-adapter} and cache-based classifiers~\cite{tip,cafo} to predict the logit bias $\mathbf{s}_{\text{bias}}$. To evaluate the effect of logit predictor, we empirically compare with several predictors, including MLP, Cache, Cache with random initialization (Cache-Random), and a simple linear probing (LP) as baseline. Specifically, we conduct experiments on ImageNet-1K by using various shots of training samples by following the same settings in \cref{Sec_Log_Fea}. All logit predictors exploit the outputs from the RN50 model of MoCov3~\cite{mocov3} as the input features. As shown in~\cref{fig:adapter}, LP achieves similar performance with Cache-Random, and both of them are clearly superior to MLP. Particularly, LP is more efficient than Cache-Random, especially for large-shot settings. The original Cache method with feature initialization achieves better performance than LP and Cache-Random, indicating that feature initialization is helpful for the logit predictor.

\begin{figure}
    \begin{subfigure}[t]{0.235\textwidth}
        \centering
        \includegraphics[width=\textwidth]{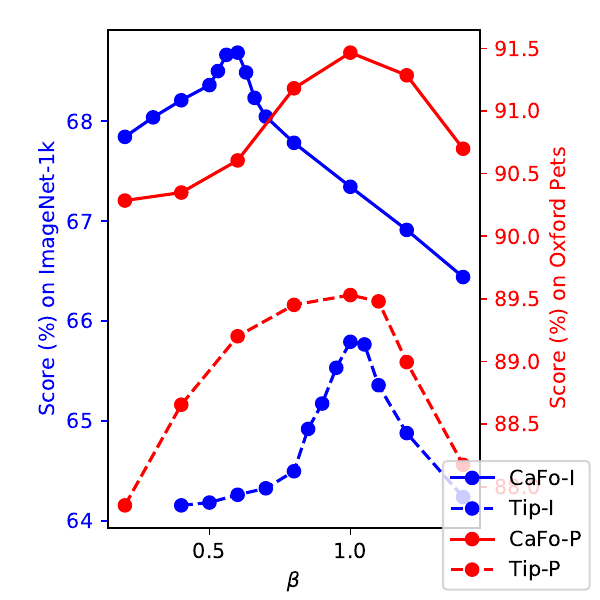}
        \caption{Results with various $\beta$}
        \label{fig:sensitivities}
    \end{subfigure}
    \hfill
    \begin{subfigure}[t]{0.235\textwidth}
        \centering
        \includegraphics[width=\textwidth]{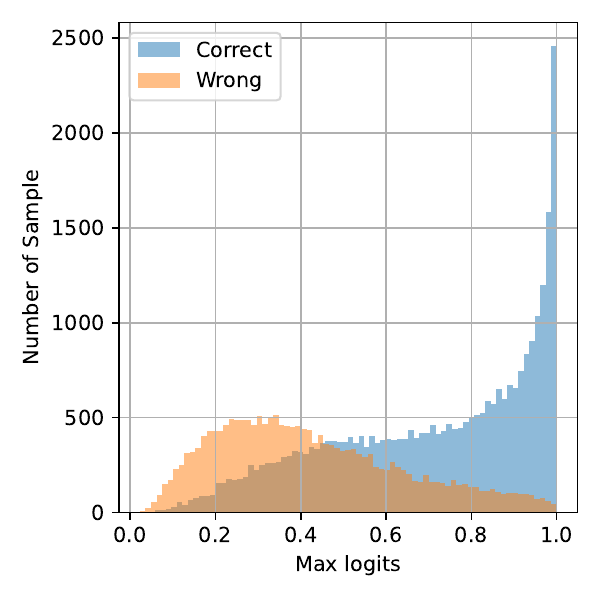}
        \caption{Visualization of max logits}
        \label{fig:logit_distribution}
    \end{subfigure} 
    \caption{(a) Results of Tip-Adapter-F and CaFo with various $\beta$ on ImageNet-1K and OxfordPets. (b) Visualization of the distribution of max logits for zero-shot CLIP on ImageNet-1K.}
\end{figure}

Furthermore, we analyze the effect of logit predictor by comparing results of the bias branch under two training strategies, i.e., individual training of bias branch and joint training of bias branch with zero-shot CLIP. Particularly, we use a simple LP as logit predictor for efficiency. The results with 16-shot training samples on ImageNet-1K are given in~\cref{tab:adaptertrain}, where we can see that the individually trained bias branch (Individual) is significantly superior to one jointly trained with zero-shot CLIP (Joint) for all logit features. Clearly, the joint training strategy makes logit bias as a pure supplement to zero-shot CLIP by considering the complementarity of auxiliary features, but it cannot fully explore the superiority of auxiliary features. Based on the above results, we conclude that \textit{feature initialization is helpful for logit predictor, while existing logit predictors do not fully explore the superiority of auxiliary features}. 

\subsubsection{Logit Fusion}\label{sec:3.2.3}

To fuse logit bias with zero-shot CLIP, a manually tuned parameter $\beta$ is used to control the effect of logit bias. As illustrated in \cref{fig:sensitivities}, we show the 16-shot results of Tip-Adapter~\cite{tip} and CaFo~\cite{cafo} with various $\beta$ on ImageNet-1K and OxfordPets datasets, where their performance is heavily affected by $\beta$, while $\beta$ is sensitive to different methods and datasets. To further analyze the effect of logit fusion, we visualize the distribution of max logits for zero-shot CLIP on ImageNet-1K. As shown in~\cref{fig:logit_distribution}, zero-shot CLIP tends to correctly classify the samples with large max logits, which indicates logits with higher confidence of zero-shot CLIP generally results in more precise classification. On the contrary, zero-shot CLIP usually mis-classifies the samples with logits of lower confidence. Therefore, we can increase effect of logit bias (i.e., value of $\beta$) for samples with low-confidence predictions of zero-shot CLIP. Based on above results, we conclude that \textit{trade-off parameter greatly affects performance of fusion, while prediction confidence of zero-shot CLIP can be regarded as an indicator of logit fusion}.

\subsection{Learning Bias via AMU}  

Based on above analysis, we propose a novel AMU-Tuning method to learn effective logit bias based on appropriate auxiliary features, effective logit predictor and adaptive logit fusion. Particularly, the overview of our AMU-Tuning is illustrated in \cref{fig:mainfig}, whose details are given as follows.

\paragraph{Auxiliary Features $\mathbf{f}^{\text{Aux}}$} According to the conclusion in \cref{Sec_Log_Fea}, we can seek the optimal auxiliary features $\mathbf{f}^{\text{Aux}}$ from a group of feature candidates based on the metrics of superiority and complementarity (\cref{eq:5}). Specifically, we employ a certain of features lying in $\Omega_S^{\text{Top-K}} \cap \Omega_C^{\text{Top-M}}$ for various downstream tasks, where $\Omega_S^{\text{Top-K}}$ and $\Omega_C^{\text{Top-M}}$ indicate the sets of  features with Top-K superiority and Top-M complementarity, respectively. For efficiency, we adopt MoCov3 model with the backbone of RN50 to obtain the auxiliary features $\mathbf{f}^{\text{Aux}}$ with no special declaration.

\paragraph{Multi-branch Training of Feature-initialized (MTFi) Logit Predictor }

As shown in ~\cref{sec_log_pre}, LP is more efficient than cache-based predictor, but the latter achieves better performance with the help of feature initialization. It encourages us to propose a feature-initialized LP for improving efficiency and effectiveness of logit predictor. Specifically, under $C$-way-$N$-shot setting with $C$ classes and $N$ samples of each class, we initialize the weights $\widehat{\mathbf{W}}$ of LP by using the mean of auxiliary features from different classes:  \vspace{-2mm}
\begin{align}\label{equ_fi}
\widehat{\mathbf{W}}_{0} & = [\mathbf{m}_1, \mathbf{m}_2,\cdots,\mathbf{m}_C]^{\mathsf{T}},  \nonumber\\
\mathbf{m}_i & = \frac{1}{N}\sum\limits_{j=1}^N \mathbf{f}_{ij}^{\text{Aux}} ,\quad i = \{1,2, \cdots ,C\},
 \vspace{-2mm}
\end{align}
where $\widehat{\mathbf{W}}_{0}$ is the initialization of $\widehat{\mathbf{W}}$, and $\mathbf{f}_{ij}^{\text{Aux}}$ is the $j$-th feature of $i$-th class. As such, our feature-initialized LP predicts logit bias $\mathbf{s}_{\text{bias}}$ for $j$-th training sample as  \vspace{-2mm}
\begin{align}
\mathbf{s}_{\text{bias}}^{j}= \widehat{\mathbf{W}}\mathbf{f}_{j}^{\text{Aux}}.
\end{align}

Furthermore, we propose a multi-branch training strategy to fully explore the superiority of auxiliary features. Specifically, besides the original classification loss (i.e., $\ell_{\text{Fusion}}$) based on the fused logit $\mathbf{s}$ , we introduce an extra training branch to minimize the cross-entropy loss between logit bias $\mathbf{s}_{\text{bias}}$ and the ground-truth label of $\mathbf{y}$ as  \vspace{-2mm}
\begin{align}\label{equ_loss}
\ell_{\text{Aux}} =  - \sum\limits_{j = 1}^{C \times N} {\mathbf{y}_j}  \cdot \log (g(\mathbf{s}_{\text{bias}}^{j})),
\end{align}
where $g(\cdot)$ is a softmax function. As such, the total loss of our multi-branch training can be formulated as:  \vspace{-2mm}
\begin{align}\label{equ_tol_loss}
\ell_{\text{total}} =(1-\lambda)\ell_{\text{Aux}} +  \lambda\ell_{\text{Fusion}},
\end{align}
where $\lambda$ is a hyper-parameter to balance effect of $\ell_{\text{Fusion}}$ and $\ell_{\text{Aux}}$. From \cref{equ_fi} and \cref{equ_loss}, we can see than our proposed MTFi logit predictor benifits feature initialization in a more efficient way, while exploring the superiority of auxiliary features by introducing an extra training branch $\ell_{\text{Aux}}$. 

\paragraph{Uncertainty-based Fusion} Based on the analysis in \cref{sec:3.2.3}, the hyper-parameter $\beta$ of bias fusion is very sensitive to models and datasets. Meanwhile, such hyper-parameter is related to  prediction confidence of zero-shot CLIP. Therefore, we present an uncertainty-based fusion to adaptively combine zero-shot CLIP with logit bias based on prediction confidence of zero-shot CLIP. Specifically, we introduce an uncertainty ($\kappa$) based on Kurtosis (i.e., the fourth moment)~\cite{confidence} to represent prediction confidence as  \vspace{-2mm}
\begin{align}\label{equ_fusion}
 \kappa  = \mathbb{E}\left[{\left(\frac{\mathbf{s}_0 - \mu}{\sigma}\right)^4}\right]^{\rho},   
\end{align}
where $\mu$ and $\sigma$ are the mean and the standard deviation of $\mathbf{s}_0$, respectively. $\rho$ is a parameter to control the power of uncertainty. As such, we can adopt $\kappa$ to balance effect of logit bias. Specifically, we increase effect of logit bias for small $\kappa$; otherwise, effect of logit bias is decreased. In conclusion, our AMU-Tuning method can be formulated as  \vspace{-2mm}
\begin{align}
 \mathbf{s}  = \mathbf{s}_0 + \frac{\beta}{\kappa}\widehat{\mathbf{W}}\mathbf{f}^{\text{Aux}},   
\end{align}
where only a lightweight LP with the parameters of $\widehat{\mathbf{W}}$ is optimized by the loss $\ell_{\text{total}}$ (\cref{equ_tol_loss}).

\section{Experiments}\label{sec:exam}
Here, we first describe implementation details of our AMU-Tuning, and then compare with state-of-the-arts (SOTA) on downstream tasks and out-of-distribution (OOD) benchmarks. Finally, we conduct ablation study on ImageNet-1K.

\subsection{Implementation Details}

Following previous works~\cite{tip,cafo}, we evaluate the effectiveness of our AMU-Tuning with [1, 2, 4, 8, 16]-shot training samples on eleven downstream tasks, including ImageNet-1K~\cite{imagenet}, StandfordCars~\cite{car}, Caltech101~\cite{caltech101}, UCF101~\cite{ucf101}, Flowers102~\cite{flower102}, Food101~\cite{food101}, DTD~\cite{dtd}, EuroSAT~\cite{eurosat}, FGVCAircraft~\cite{fvga}, OxfordPets~\cite{pet}, and SUN397~\cite{sun397}. Specifically, we trained our AMU-Tuning model (i.e., a lightweight LP) on all downstream tasks within 50 epochs, where the AdamW optimizer is used with the initial learning rate of 0.001 and batch size of 8. The hyper-parameters of $\lambda$ in \cref{equ_loss} and $\rho$ in \cref{equ_fusion} are decided by cross-validation on the validation sets. All program is implemented by PyTorch~\cite{pytorch}/
 MindSpore and run on a single RTX 3090 GPU. Source code is available at \url{https://github.com/TJU-sjyj/AMU-Tuning}.

\begin{figure*}
    \centering
    \includegraphics[width=\textwidth]{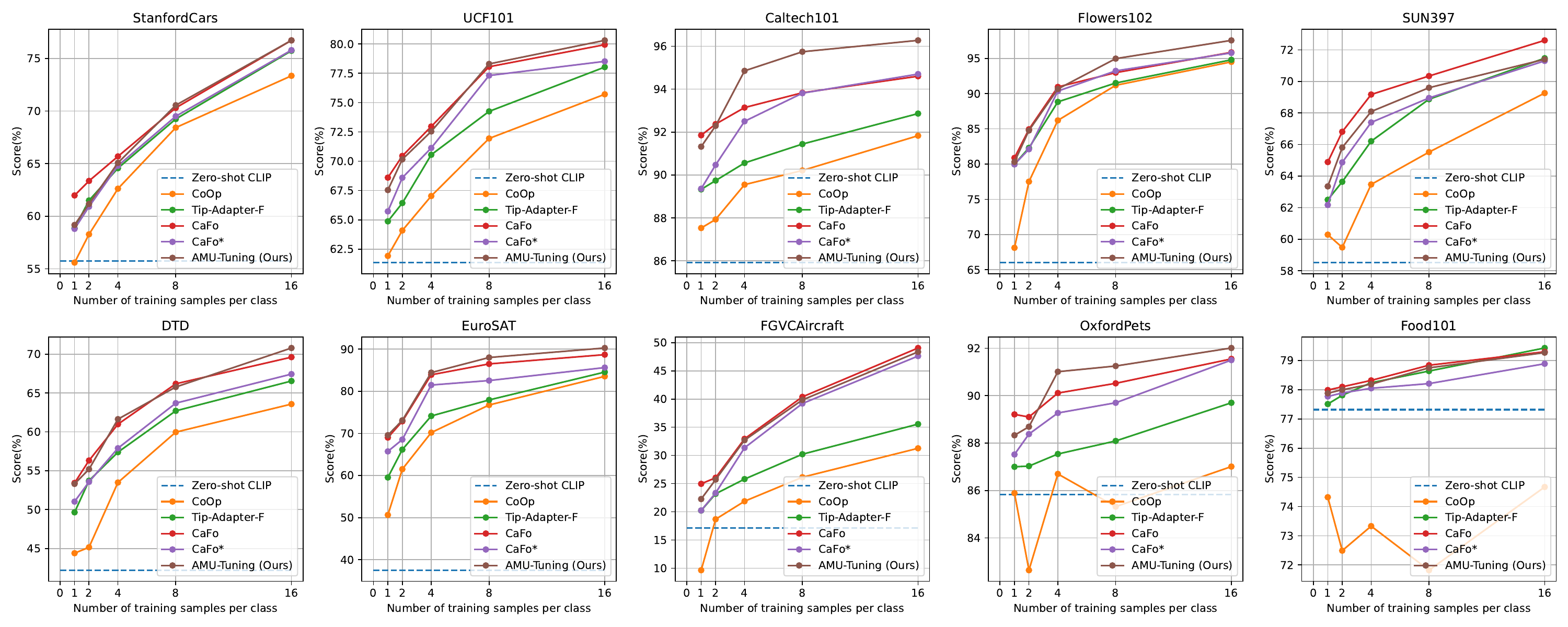}
    \caption{Comparison (in \%) of different SOTA methods under various few-shot settings on ten downstream tasks.}
    \label{fig:10dataset}
\end{figure*}

\begin{figure}
    \centering
    \includegraphics[width=0.33\textwidth]{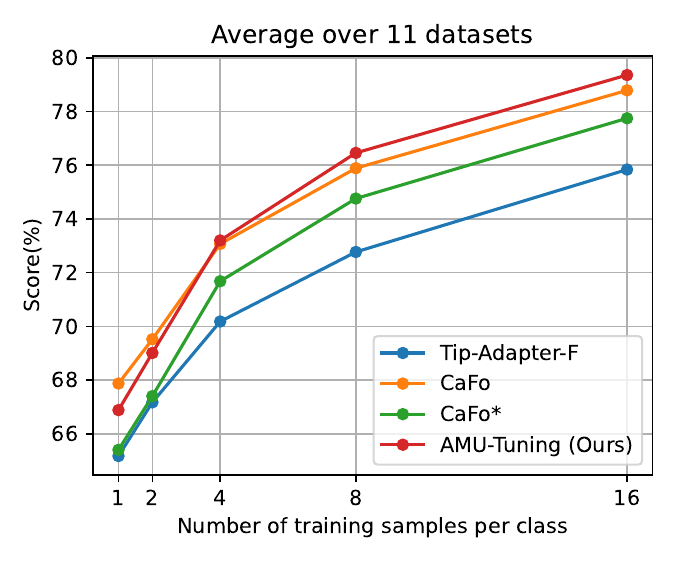}
    \caption{Results on eleven downstream tasks by average.}
    \label{fig:avg}
\end{figure}

\begin{table}[t]
  \tabcolsep=2pt 
  \centering
    \begin{tabular}{lccccc}
    \toprule
    \multirow{2}[4]{*}{Method}& \multicolumn{5}{c}{Score} \\
\cmidrule{2-6}          & 1-shot     & 2-shot     & 4-shot     & 8-shot     & 16-shot \\
    \midrule
    LP-CLIP~\cite{clip} & 22.17  & 31.90  & 41.20  & 49.52  & 56.13  \\
    CoOp~\cite{coop}  & 57.15  & 57.81  & 59.99  & 61.56  & 62.95  \\
    CLIP-Adapter~\cite{clip-adapter} & 61.20  & 61.52  & 61.84  & 62.68  & 63.59  \\
    VT-CLIP~\cite{vt-clip} & 60.53  & 61.29  & 62.02  & 62.81  & 63.92  \\
    Tip-Adapter-F~\cite{tip} & 61.32  & 61.69  & 62.52  & 64.00  & 65.51  \\
    CaFo~\cite{cafo}  & \textbf{63.80}  & \textbf{64.34}  & \underline{65.64}  & \underline{66.86}  & \underline{68.79}  \\
    CaFo$^{\star}$~\cite{cafo} & 61.58  & 62.76  & 64.31  & 66.25  & 68.05  \\
    \midrule
    AMU-Tuning (Ours)  & \underline{62.60}  & \underline{64.25}  & \textbf{65.92}  & \textbf{68.25}  & \textbf{70.02 } \\
    \bottomrule
    \end{tabular}%
    \caption{Comparison (in \%) of different SOTA methods on ImageNet-1K under various few-shot settings.}
  \label{tab:imagenet}%
\end{table}%

\begin{table}[t]
 \tabcolsep=1.8pt
  \centering
    \begin{tabular}{lccccc}
    \toprule
    \multirow{2}[4]{*}{Dataset} & Source & \multicolumn{4}{c}{Target} \\
\cmidrule(lr){2-2} \cmidrule(lr){3-6}          & IN-1K & v2    & -S & -A    & -R \\
    \midrule
    ZS-CLIP~\cite{clip} & 60.33  & 53.27  & 35.44  & 21.65  & 56.00  \\
    CoOp~\cite{coop}  & 62.95  & 55.40  & 34.67  & \underline{23.06}  & \underline{56.60}  \\
    CLIP-Adapter~\cite{clip-adapter} & 63.59  & 55.69  & 35.68  & -      &-  \\
    Tip-Adapter-F~\cite{tip} & 65.51  & 57.11  & 36.00  &-       &-  \\
    CaFo~\cite{cafo}  & \underline{68.79}  & \underline{57.99}  & \underline{39.43}  &-       &-  \\
    AMU-Tuning (RN50)  & \textbf{70.02}  & \textbf{58.64}  & \textbf{40.04}  &\textbf{25.65}    &\textbf{57.10}  \\
    \midrule
    CoCoOp~\cite{cocoop}  & 71.02& 64.20& 47.99&49.71&75.21\\
    MaPLe~\cite{maple}  & 70.72& 64.07& 49.15&50.90&76.98\\
    AMU-Tuning (ViT)  & \textbf{74.98}& \textbf{65.42}& \textbf{50.37}&\textbf{52.05}&\textbf{78.09}\\
    \bottomrule
    \end{tabular}%
    \caption{Comparison (\%) of different methods under OOD setting.}
  \label{tab:Shift}
\end{table}
\vspace{-2mm}
\subsection{Comparison with SOTA Methods}
\paragraph{Results on Downstream Tasks.} To verify the effectiveness of our AMU-Tuning method, we first compare with several SOTA CLIP-based few-shot classification methods with backbone of RN50 on ImageNet-1K, including CoOp~\cite{coop}, CLIP-adapter~\cite{clip-adapter}, VT-CLIP~\cite{vt-clip}, Tip-Adapter-F~\cite{tip} and CaFo~\cite{cafo}. Particularly, we implement a CaFo variant (namely CaFo$^{\star}$) by excluding use of the extra DALL-E and GPT-3. The results are shown in \cref{tab:imagenet}, where our AMU-Tuning clearly outperforms all SOTA methods (except CaFo) under various few-shot settings. Comparing with CaFo, our AMU-Tuning achieves better performance under larger-shot settings (i.e., $>$ 4-shot). For smaller-shot settings  (i.e., 1-shot and 2-shot), our AMU-Tuning is slightly inferior to CaFo. The reason behind this phenomenon lies in that CaFo uses the extra training samples generated by DALL-E, which contributes to large gains for smaller-shot settings. Comparing with CaFo$^{\star}$, our AMU-Tuning achieves 1.02\%, 1.49\%, 1.61\%, 1.39\% and 1.97\% for [1, 2, 4, 8, 16]-shot training samples, respectively. Furthermore, we compare with SOTA methods on ten additional downstream tasks, which cover various scenarios, e.g., objects, satellite photos, textures, and scene images. As illustrated in~\cref{fig:10dataset}, our AMU-Tuning achieves the best results on most of downstream tasks. Meanwhile,~\cref{fig:avg} shows AMU-Tuning clearly outperforms Tip-Adapter~\cite{tip} and CaFo$^{\star}$ on average over eleven downstream tasks. These results above clearly demonstrate the effectiveness and strong generalization of our AMU-Tuning.


\begin{table}[t]
  \centering
    \begin{tabular}{ccc|ccc}
    \toprule
    \multicolumn{3}{c|}{Component} & \multicolumn{3}{c}{Score (\%)} \\
    \midrule
    \multicolumn{1}{c}{AUX} & \multicolumn{1}{c}{MTFi} & \multicolumn{1}{c|}{UF} & 1-shot     & 4-shot     & 16-shot \\
    \midrule
   \multicolumn{3}{c|}{Baseline} & 61.16  & 62.33  & 65.34  \\
    \midrule
     \checkmark &       &       & 62.15  & 65.31  & 69.35  \\
     & \checkmark &       & 61.83  & 63.16  & 66.17\\
     &       & \checkmark & 61.70  & 63.08  & 65.90\\
    \checkmark & \checkmark &       & 62.35  & 65.61  & 69.72  \\
    \checkmark & \checkmark & \checkmark & \textbf{62.60}& \textbf{65.92}& \textbf{70.02}\\
    \bottomrule
    \end{tabular}%
\caption{Results of AMU-Tuning with various modules on IN-1K.}
\label{tab:module}
\end{table}
\vspace{-2mm}
\paragraph{Robustness to OOD.}Following previous works~\cite{coop,cocoop}, we further verify the robustness of AMU-Tuning to OOD benchmarks. Specifically, we directly adopt the models fine-tuned on ImageNet-1K (IN-1K) with 16-shot training samples to four OOD benchmarks, including ImageNet-V2~\cite{imagenet-v2}, ImageNet-Sketch~\cite{imagenet-s}, ImageNet-A~\cite{imagenet-a}, and ImageNet-R~\cite{imagenet-r}. As shown in~\cref{tab:Shift}, our AMU-Tuning achieves the best results on all OOD benchmarks, which can transfer performance gains on IN-1K to OOD setting. Since  CLIP models are trained on a mass of external image-text pairs, potentially suffering from the risk of data leakage. However, previous works~\cite{clip,quality} show  the overlapping data brings little effect on performance, perhaps the self-supervised settings (without ground-truth labels of  samples). Besides, all of our compared methods are built upon CLIP model, these results above verify that our AMU-Tuning is robust to OOD setting.

 \vspace{-2mm}
\begin{table}[t]
\tabcolsep=3pt
  \centering
    \begin{tabular}{lcccc}
    \toprule
    \multicolumn{1}{l}{\multirow{2}[4]{*}{Models}} & \multicolumn{4}{c}{Backbone} \\
\cmidrule{2-5}          & RN50  & RN101 & ViT-B/32 & ViT-B/16 \\
    \midrule
    ZS-CLIP~\cite{clip} & 60.33 & 65.53 & 63.80  & 68.73 \\
    CoOp~\cite{coop} & 62.95 & 66.60  & 66.85 & 71.92 \\
    CLIP-Adapter~\cite{clip-adapter}& 63.59 & 65.39 & 66.19 & 71.13 \\
    Tip-Adapter-F~\cite{tip}& 65.51 & 68.56 & 68.65 & 73.69 \\
    CaFo~\cite{cafo} & 68.79 & 70.82 & 70.82 & 74.48 \\
    CaFo$^{\star}$~\cite{cafo} & 68.03 & 70.21 & 70.44 & 74.11 \\
    AMU-Tuning (Ours) & \textbf{70.02} & \textbf{71.58} & \textbf{71.65} & \textbf{74.98} \\
    \midrule    
    \end{tabular}%
      \caption{Comparison (\%) of SOTA methods with different visual encoders of CLIP on IN-1K with 16-shot training samples.}
  \label{tab:encoder}%
\end{table}%

\subsection{Ablation study}
Our AMU-Tuning method involves three key components, i.e., appropriate auxiliary features (AUX), multi-branch training feature-initialized (MTFi) logit predictor, and uncertainty-based fusion (UF). To evaluate effect of different modules on AMU-Tuning, we conduct ablation studies on ImageNet-1K (IN-1K) dataset. Specifically, \cref{tab:module} shows the results of AMU-Tuning with various components, where the baseline is built based on auxiliary features of CLIP with simple LP following by $\beta$-fusion. From \cref{tab:module} we can see that all components bring clear performance gains over the baseline. Besides, going beyond our strong baseline with the auxiliary features of MoCov3, MTFi and UF modules further bring 0.5\%$\sim$0.7\% gains, while achieving SOTA performance. 

Furthermore, we compare AMU-Tuning with SOTA methods by using various visual encoders of CLIP, including RN50, RN101, ViT-B/32 and ViT-B/16. Note that our AMU-Tuning utilizes a RN50 pre-trained by MoCov3 to obtain the auxiliary features. As listed in~\cref{tab:encoder}, AMU-Tuning consistently outperforms all compared methods for different visual encoders of CLIP. Particularly, although existing methods (e.g., Tip-Adapter and CaFo) employ stronger visual encoders (e.g., ViT-B) to compute logit bias, AMU-Tuning can obtain better results by using RN50-based auxiliary features. These results above verify strong generalization of our AMU-Tuning again.

\vspace{-2mm}
\section{Conclusion}

To better understand the effectiveness of existing CLIP-based few-shot learning methods, this work first introduces a unified formulation from the perspective of logit bias. Then,  we disassemble computation of logit bias into three key components, i.e., logit features, logit predictor, and logit fusion, and analyze the effect on performance of few-shot classification. Furthermore, our empirical analysis encourages us to propose AMU-Tuning method for effective CLIP-based few-shot classification, which learns logit bias by exploiting the appropriate auxiliary features with multi-branch training feature-initialized LP following by an uncertainty-based fusion. Extensive experiments on both downstream tasks and out-of-distribution datasets demonstrate the effectiveness of our method. We hope our analysis from the perspective of logit bias could provide a new insight for CLIP-based few-shot learning, while encouraging more effective logit bias learning methods.

{
    \small
    \bibliographystyle{ieeenat_fullname}
    \bibliography{main}
}

\clearpage
\maketitlesupplementary
\maketitle

\renewcommand{\thetable}{S\arabic{table}}
\renewcommand{\thefigure}{S\arabic{figure}}
\renewcommand{\thesection}{S\arabic{section}}
\renewcommand{\theequation}{S\arabic{equation}}
\renewcommand{\thefootnote}

\setcounter{section}{0}
\setcounter{table}{0}
\setcounter{equation}{0}
\setcounter{figure}{0}
In the supplementary material, we conduct more experiments to further investigate the effectiveness of our AMU-Tuning method. Specifically, we first analyze the computation complexity of AMU-Tuning, and then compare the MTFi predictor with LP by using various backbone models. Subsequently, we conduct the analysis on the effect of hyper-parameters $\lambda$ and $\rho$ in the AMU-Tuning method. Finally, we compare different methods to compute the confidence $\kappa$ in uncertainty fusion of Eq.\ (11). Note that all experiments are conducted on ImageNet-1K.

\section{Analysis on Computation Complexity.}In this section, we compare AMU-Tuning with its counterparts in terms of computation complexity on a single RTX 3090 GPU, including training time over 500 steps with batch size of 16384 and number of trainable parameters. \cref{tab:complex} gives the results under 16-shot setting, where we can see that AMU-Tuning has the fastest training speed and the fewest trainable parameters, since AMU-Tuning only optimizes a lightweight LP. These results verify the efficiency of our AMU-Tuning.

\begin{table}[htbp]
\centering
\begin{tabular}{lcc}
\toprule
 Model &  Time (s) $\downarrow$ & Params. (M)$\downarrow$\\
 \midrule
 Tip-Adapter-F~\cite{tip}& 22.30 & 16.38\\
 CaFo~\cite{cafo} & 87.88 & 49.15 \\ 
 AMU-Tuning (Ours) & \textbf{6.63}& \textbf{2.05}\\
 \bottomrule
 \end{tabular}%
 \caption{Computational complexity of different methods in terms of training time (Time) and trainable parameters (Params.). }
 \label{tab:complex}
 \end{table}%

\section{Comparison of MTFi with LP Using Different Backbones}

In this section, we conduct an in-depth analysis using extra backbones (i.e., DINO~\cite{dino} and MAE~\cite{mae}) to evaluate the effectiveness of MTFi on ImageNet-1K~\cite{imagenet}. Specifically, \cref{tab:dino_result} respectively shows the results of linear probing (LP) and our MTFi predictors with auxiliary features of DINO and MAE, while apparent that the implementation of MTFi leads to a remarkable boost in the model's performance.

\begin{table}[htbp]
\centering
\begin{tabular}{lccc}
\toprule
Method &  1-shot & 4-shot & 16-shot \\
\midrule
MAE~\cite{mae}+LP & 61.20 & 62.76 & 65.49\\
MAE + MTFi & \textbf{61.44} & \textbf{63.32} & \textbf{65.99}\\
\midrule
DINO~\cite{dino}+LP & 61.62 & 64.43 & 68.32\\
DINO+MTFi & \textbf{61.97}& \textbf{65.05} & \textbf{69.21}\\
\bottomrule
\end{tabular}%
\caption{Comparison(\%) of MTFi in extra backbones.}
\label{tab:dino_result}
\end{table}%

\begin{figure}[h]
    \begin{subfigure}[t]{0.235\textwidth}
        \centering
        \includegraphics[width=\textwidth]{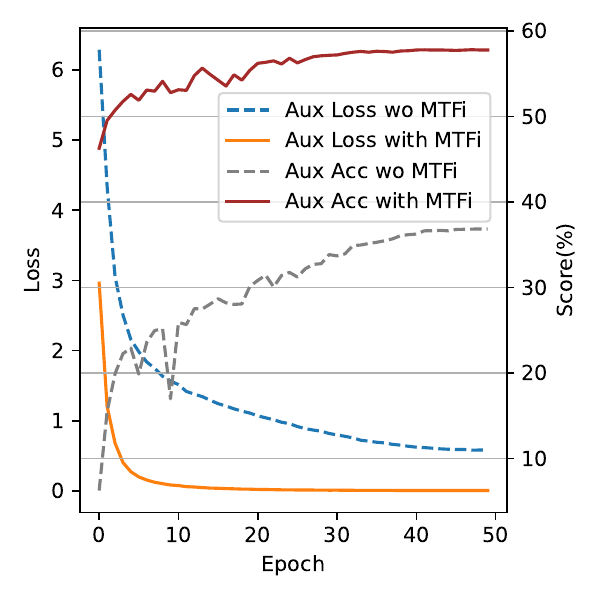}
        \caption{Loss and accuracy of auxiliary branch}
        \label{fig:loss_compare_1}
    \end{subfigure}
    \hfill
    \begin{subfigure}[t]{0.235\textwidth}
        \centering
        \includegraphics[width=\textwidth]{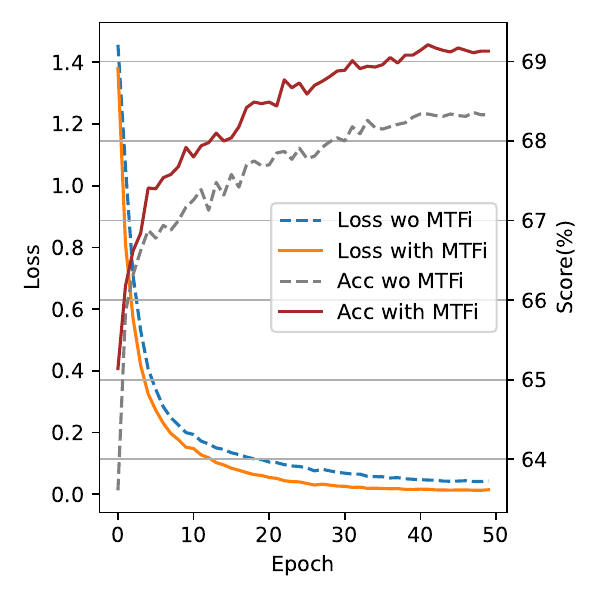}
        \caption{Loss and accuracy overall model}
        \label{fig:loss_compare_2}
    \end{subfigure}
    \caption{Comparison of loss and accuracy curves with/without the MTFi method. (a) is auxiliary loss and accuracy curve,while (b) is total loss and accuracy curve.}
    \label{fig:loss_compare}
\end{figure}

Furthermore, we visualize the loss convergence curves for the DINO mode with MTFi trained in the 16-shot setting. As shown in \cref{fig:loss_compare} both $\ell_{\text{total}}$ and $\ell_{\text{Aux}}$ convergence faster, whlie the individual accuracy (\%) of the auxiliary branch increases from 36.24 to 57.79, and the overall model accuracy (\%) improves from 68.32 to 69.21. The above results demonstrate that our MTFi can achieve significant performance improvement while benefitting higher training efficiency (e.g., DINO).

\section{Effect of Parameter $\lambda$}
\begin{figure}[h]
    \centering
    \includegraphics[width=0.45\textwidth]{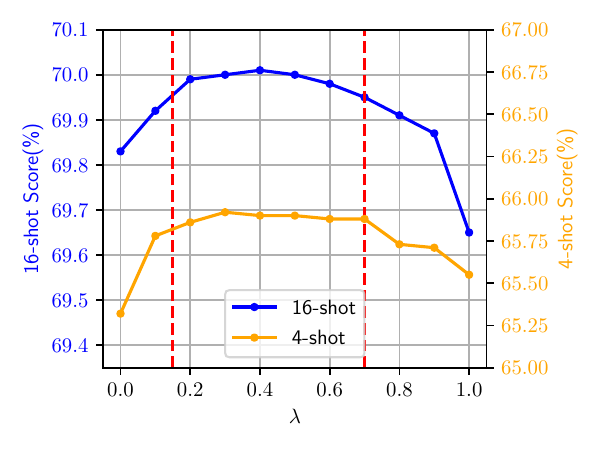}
    \caption{Comparison on AMU-Tuning with different $\lambda$.}
    \label{fig:lambda}
\end{figure}

In Eq.\ (9), we introduce a hyper-parameter $\lambda$ that balance the effect between $\ell_{\text{Aux}}$ and $\ell_{\text{Fusion}}$. In \cref{fig:lambda}, we examine the performance of the AMU-Tuning method with 4-shot and 16-shot on ImageNet-1K under various values of $\lambda$. It is evident that the performance of AMU-Tuning method remains stable when $\lambda$ is in the range of 0.2 to 0.6, with fluctuations below 0.05\%. Therefore, we generally set $\lambda$ to $0.4$ throughout our experiments.

\section{Effect of Parameter $\rho$}
\begin{figure}[h]
    \centering
    \includegraphics[width=0.45\textwidth]{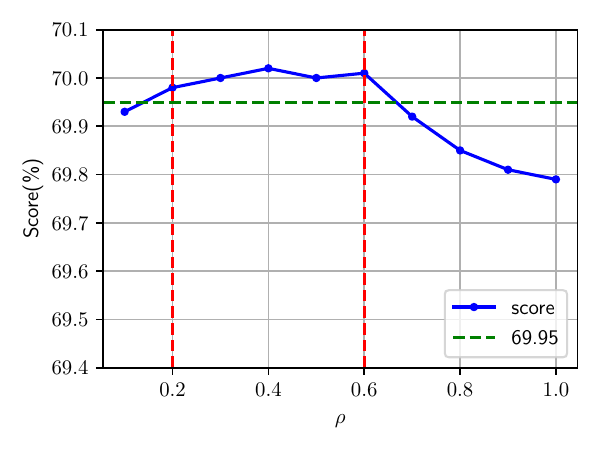}
    \caption{Comparison on AMU-Tuning with different $\rho$.}
    \label{fig:rho}
\end{figure}

In Eq.\ (10), we introduce a hyper-parameter $\rho$  to control the power of uncertainty. In~\cref{fig:rho}, we examine the performance of the AMU model with 16-shot on ImageNet-1K under various values of $\rho$. It is evident that the performance of AMU performance stable when $\rho$ is in the range of 0.2 to 0.6. In our experiments, we generally set $\rho$ to 0.4.

\section{Comparison of Different Uncertainty-based Fusion Methods}

Accroding to the observation in Sec. 3.2.3, we have that the largest logit value of zero-shot CLIP is consistently high, when the sample is correctly classified. Therefore, we devise various approaches for calculating the confidence parameter $\kappa$. First, we compute the confidence score $\kappa$ based on the largest logit value directly which formularized as
\begin{equation}\label{eqmax}
{\kappa _{\rm{Max}}} = Max1({{\mathbf{s}}_{\rm{0}}})^{\rho},
\end{equation}
where $Max1(\cdot)$ computes the largest logit value. Furthermore, we also design a method that simultaneously utilizes the largest and second largest logit values, as suggested in~\cite{confidence}. The formula is as follows:
\begin{equation}\label{eqtp2}
{\kappa _{{\rm{Top2}}}} = \left(\frac{Max1({{\mathbf{s}}_0})- Max2(\mathbf{s}_{\rm{0}})}{abs(Max1(\mathbf{s}_{\rm{0}})+Max2(\mathbf{s}_{\rm{0}}))}\right)^{\rho},
\end{equation}
$Max2(\cdot)$ computes the second largest logit value. We also explore a confidence calculation method based on energy~\cite{energy}, for $C$ classes $\kappa$ formulated as follows:
\begin{equation}\label{eqenergy}
{\kappa _{{\rm{Energy}}}} = \left(\log \sum\limits_{i = 1}^C {{e^{{\mathbf{s}}_0^i}}}  \right)^{\rho}.
\end{equation}

\begin{figure}
    \centering
    \includegraphics[width=0.45\textwidth]{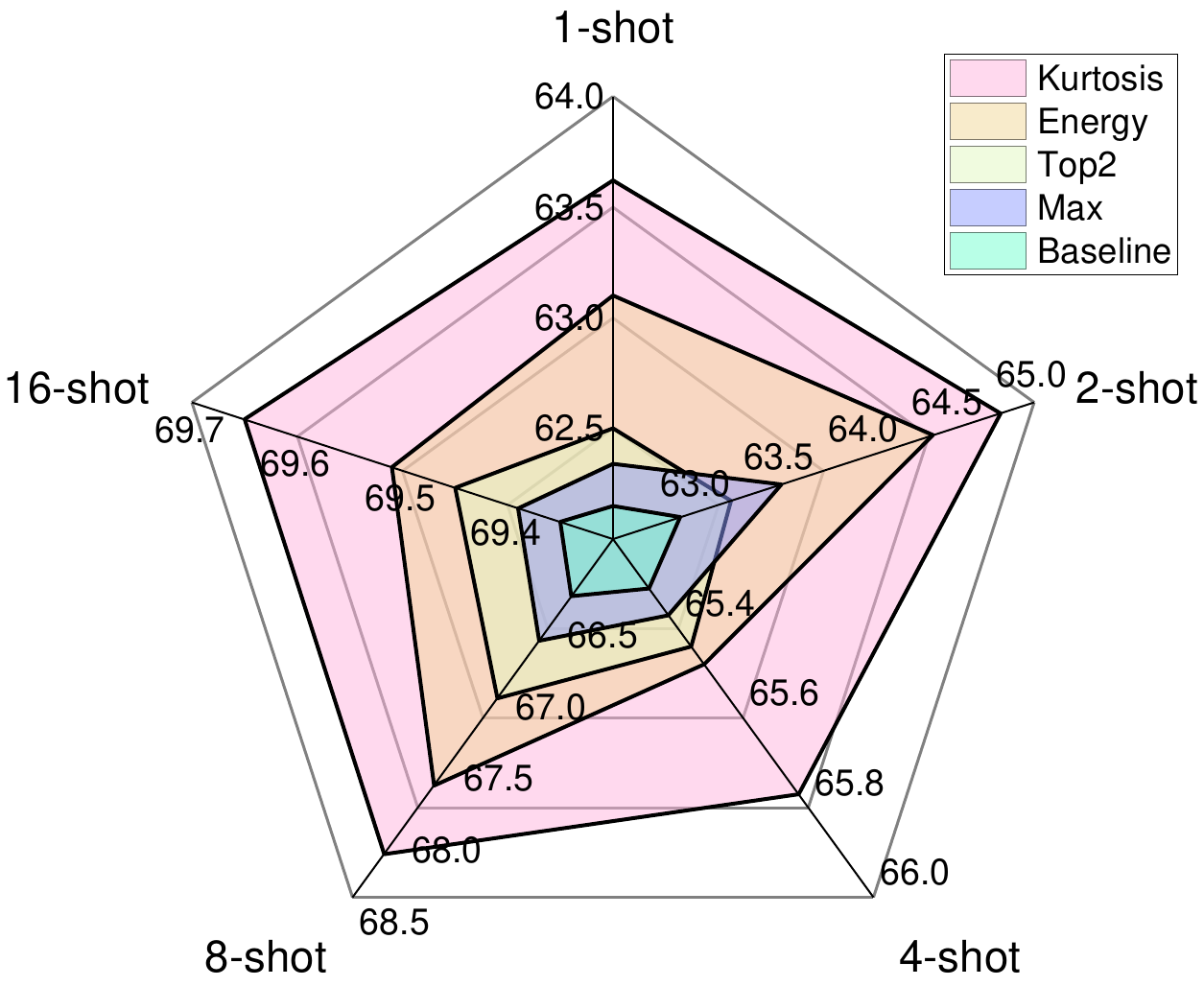}
    \caption{Comparison(\%) of different confidence calculation methods.}
    \label{fig:um}
\end{figure}
We compare different methods in \cref{eqmax,eqtp2,eqenergy} and our kurtosis-based confidence in Eq.\ (10) on ImageNet-1K with the auxiliary features of MoCov3~\cite{mocov3}. As present in~\cref{fig:um} several confidence computation methods can lead to performance improvement while our kurtosis-based approach achieves the better performance than other compared methods for all cases.



\end{document}